\documentclass[letterpaper, 10 pt, conference]{ieeeconf}  
\IEEEoverridecommandlockouts                              
\usepackage{cite,xspace,color}
\usepackage{amsmath,amssymb,amsfonts}
\usepackage{algorithmicx}
\usepackage{graphicx}
\usepackage{textcomp}
\usepackage{algorithm}
\usepackage[noend]{algpseudocode}
\def\BibTeX{{\rm B\kern-.05em{\sc i\kern-.025em b}\kern-.08em
    T\kern-.1667em\lower.7ex\hbox{E}\kern-.125emX}}
\usepackage{tabto}
\usepackage{dblfloatfix}
\usepackage{url}

\def\lar{\leftarrow}
\def\ba{\begin{array}}
\def\ea{\end{array}}
\def\beq{\begin{equation}}
\def\eeq#1{\label{#1}\end{equation}}
\def\beqq{\begin{equation*}}
\def\eeqq{\end{equation*}}
\def\no{\ii{not}}
\def\ii#1{\hbox{\it #1\/}}

\def\lar{\leftarrow}
\def\ba{\begin{array}}
\def\ea{\end{array}}
\def\beq{\begin{equation}}
\def\eeq#1{\label{#1}\end{equation}}
\def\beqq{\begin{equation*}}
\def\eeqq{\end{equation*}}
\def\no{\ii{not}}
\def\ii#1{\hbox{\it #1\/}}

\newcommand{\amp}[1]{\ensuremath{\text{\textsl{{\&}}}\!\,\mathit{#1}}}
\newcommand{\ext}[3]{\ensuremath{\amp{#1}[#2](#3)}}

\newcommand{\dlvhex}{\textsc{dlvhex}\xspace}

\newtheorem{prop}{\noindent \textbf{Proposition}}

\title{\LARGE \bf  A Formal Framework for Robot Construction Problems: \\
A Hybrid Planning Approach}
\author{Faseeh Ahmad  \and Esra Erdem \and Volkan Patoglu
\thanks{F. Ahmad,  E. Erdem and V. Patoglu are with the Faculty of Engineering and Natural Sciences at Sabanc\i\ University, \.Istanbul, Turkey.
        {\tt\footnotesize \{faseehahmad,esraerdem,vpatoglu\}@sabanciuniv.edu}}%
}

\begin{document}

\maketitle
\thispagestyle{empty}
\pagestyle{empty}

{
\begin{abstract}
 We study robot construction problems where multiple autonomous robots rearrange stacks of prefabricated blocks to build stable structures. These problems are challenging due to ramifications of actions, true concurrency, and requirements of supportedness of blocks by other blocks and stability of the structure at all times. We propose a formal hybrid planning framework to solve a wide range of robot construction problems, based on Answer Set Programming. This framework not only decides for a stable final configuration of the structure, but also computes the order of manipulation tasks for multiple autonomous robots to build the structure from an initial configuration, while simultaneously ensuring the stability, supportedness and other desired properties of the partial construction at each step of the plan. We prove the soundness and completeness of our formal method with respect to these properties. We introduce a set of challenging robot construction benchmark instances, including bridge building and stack overhanging scenarios, discuss the usefulness of our framework over these instances, and demonstrate the applicability of our method using a bimanual Baxter robot.
\end{abstract}
}


\section{Introduction}

\begin{figure*}[h!]
	\centering
    \resizebox{1.85\columnwidth}{!}{\includegraphics{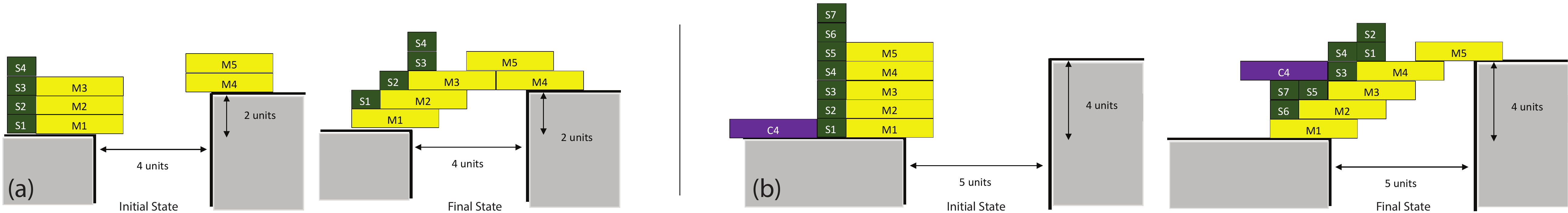}}
    \vspace{-.75\baselineskip}
	\caption{Stable construction of unlevel bridges}
    \vspace{-0.5\baselineskip}
	\label{fig:unlevel-bridges}
\end{figure*}
\begin{figure*}[h!]
	\centering
    \resizebox{1.85\columnwidth}{!}{\includegraphics{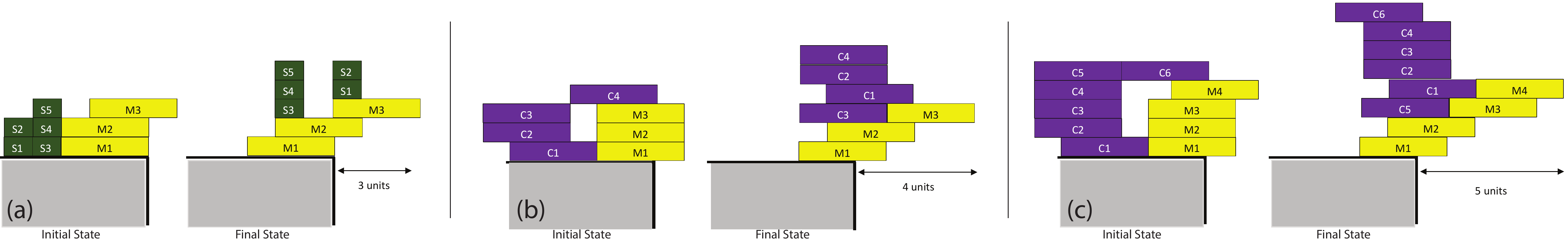}}
    \vspace{-.75\baselineskip}
	\caption{Stable construction of overhangs}
    \vspace{-0.5\baselineskip}
	\label{fig:overhangs}
\end{figure*}
\begin{figure*}[h!]
	\centering
    \resizebox{1.85\columnwidth}{!}{\includegraphics{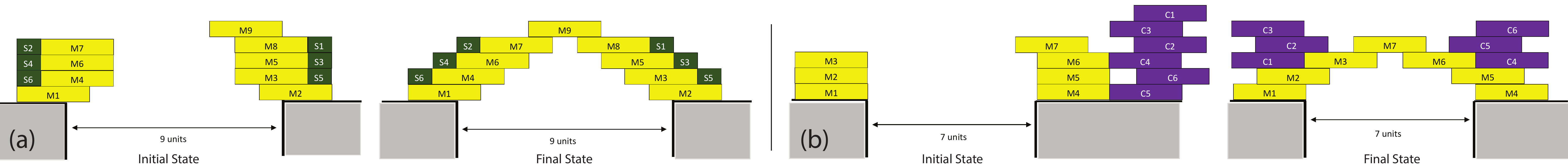}}
    \vspace{-.75\baselineskip}
	\caption{Stable construction of symmetric bridges}
    \vspace{-.75\baselineskip}
	\label{fig:bridges}
\end{figure*}

The construction industry relies on manual labor as its primary source of productivity, while robots promise to dramatically improve the speed and quality of construction work by automating the repetitive and labor intensive tasks~\cite{bock2008}. Even though automation can improve the efficiency and the productivity of certain construction tasks, in these approaches, deciding for the design of the structure to be built, planning of the robot motions, and proper ordering of robot actions still need to be decided manually. Robotics will have a major impact on the construction industry, if these reasoning tasks can also be performed automatically. For instance, it would be very beneficial if a group of autonomous search and rescue robots could automatically build bridges in a disaster zone, by rearranging stacks of prefabricated building materials that are accessible to them.

Consider, for instance, the robot construction problems in Figure~\ref{fig:unlevel-bridges} specified by their initial states; in each problem, the goal is to build a stable bridge. The construction area is limited, and an upper bound is given for the number of actions. The problem shown in Figure~\ref{fig:unlevel-bridges}(b) further requires construction to start from the left side and proceed towards the right side. The solutions to these problems require finding stable goal configurations of prefabricated blocks so that they connect the two sides (e.g., the final states in Figure~\ref{fig:unlevel-bridges}), handling ramifications of actions (e.g., when C4 is placed over~S7, it becomes on S5 as an indirect effect as in Figure~\ref{fig:unlevel-bridges}(b)), construction and incorporation of subassemblies (e.g., the subassembly of S4, S3, M3 in the final state in Figure~\ref{fig:unlevel-bridges}(a)), using blocks or subassemblies as counterweights (e.g., all the small blocks and C4 in the final state in Figure~\ref{fig:unlevel-bridges}(b)), and maintaining stability of the structure at all times. Other examples are provided to illustrate construction of overhangs (Figures~\ref{fig:overhangs} and~\ref{fig:overhang}), symmetric bridges (Figure~\ref{fig:bridges}), and other interesting structures (Figure~\ref{fig:plans}) to show further challenges, such as the need for concurrency of actions (e.g., moving S1 and S2 onto L1 at the same time as in Figure~\ref{fig:plans}(d)).

{
Sophisticated dependencies between preconditions and effects of robotic actions (due to, e.g., ramifications, concurrency, and supportedness constraints) in construction problems necessitate the use of task planners to decide the order of manipulations.} Furthermore, given the collision-free manipulation and stable configuration requirements on every step of the plan, the construction problem cannot be addressed using solely a task planner, as these feasibility checks need to be performed in continuous domains. 

We propose a formal hybrid planning framework for robot construction problems, where multiple autonomous robots rearrange stacks of prefabricated blocks to build stable structures, such as bridges. It is based on the expressive computational logic of Answer Set Programming. The planning framework decides for a stable final configuration of blocks, and computes the order of manipulation tasks for multiple robots to build it from an initial configuration. {
Furthermore, it prevents nonsensical configurations, like circular configuration of blocks (Proposition~\ref{prop:supported}), solves the ramification problem through recursive definitions of global locations of blocks from their relative locations (Proposition~\ref{prop:above}), guarantees desired properties, like the stability of a construction (Proposition~\ref{prop:stability}) and the connectedness of two sides of a bridge (Proposition~\ref{prop:connected}). In addition to these soundness results, our framework guarantees completeness by ensuring the computation of all valid construction plans subject to such properties and constraints and whose lengths are less than a given maximum makespan (Proposition~\ref{prop:complete}).}


\section{Related Work}

To the best of authors' knowledge, this is the first robotic construction study that addresses a variety of multi-robot stack rearrangement planning problems for building stable structures of different sorts. In the literature, there exist several studies that focus on different specific aspects of the robotic construction task: deciding for the stability of a given structure (e.g., from an image obtained from Angry Birds), deciding for the existence of a specified stable structure (e.g., a maximum overhang) from a given set of identical blocks or an unspecified stack from a given set of different sizes of objects (e.g., like stones), planning for towers of identical blocks (e.g.. the blocks world) ignoring stability, etc. Let us go over them to better understand the challenges of the robot construction problems that we study.

\textbf{The blocks world:} The well-known blocks world problems~\cite{WINOGRAD1972} have been widely studied by AI community; {
it is proven to be NP-complete for polynomially bounded plans~\cite{gupta1992}}. Blocks world problems are quite restricted compared to robot construction problems, since while proposing the problem, Winograd's interest was in language rather than in construction problems. For instance, the blocks world deals with identical blocks and allows a block to be placed on a flat surface or on another block, but not on multiple blocks as necessitated by the robot construction problems. It does not allow manipulation of subassemblies, use of counterweights and scaffolds, or concurrent placements of blocks, either. Also, there is no consideration of feasibility checks to ensure the stability of the stack at each step of a plan.

Later, Fahlman~\cite{fahlman1974} has introduced a set of robot construction problems where the goal is for a robot to build specified structures out of simple blocks of different shapes and sizes. These problems allow incorporation of subassemblies into the final design, and the use of extra blocks as temporary supports or counterweights during construction; they also consider collisions of blocks and instability of the structures, but not motion planning. Since Fahlman's main interest was in maximizing common sense (rather than soundness, completeness or optimality), he implemented a planning system guided with heuristics to solve some of these problems. These problems have not been investigated with a formal approach since then.

\textbf{Maximum overhang puzzle:}  Mathematicians and theoretical computer scientists have studied a classic puzzle that aims to determine the maximum overhang achievable by a stack of identical blocks~\cite{hall2005,paterson2006,paterson2009a,paterson2009b}. A relatively recent solution~\cite{paterson2009a,paterson2009b} to this 150 year old puzzle, honored with the prestigious David P. Robbins Prize in mathematics, has introduced the use of blocks as counterbalance to improve upon the well-established solution. While the maximum overhang problem focuses on the determination of a stable and optimal final configuration of identical blocks, the planning aspects of the construction problem to attain the goal configuration is not considered within the scope of these studies.

\textbf{Image understanding and qualitative reasoning in games:} Applications in scene understanding from 2D pictures and computer games require inferring physical relations among objects~\cite{gupta1992,jia2015,stephenson2016}. Determination of  stability of stacked objects and supportedness among objects have been studied, commonly with qualitative reasoning approaches~\cite{walega2016}. Determination of stable final configuration of constructions has also been studied in computer games, like Angry Bird~\cite{ferreira2014generating,stephenson2016procedural,CalimeriFGHIR0T16}. These studies focus on the physical relations of a given final configuration and do not address the block rearrangement problem to build  stable constructions. 

\textbf{Stability of assemblies:} In robotics, static stability~\cite{blum1970,livesley1978,livesley1992,schimmels1994,mattikalli1996,zwick2002,whiting2012,mojtahedzadeh2015} and dynamic stability~\cite{pang2000,spanos2001} of assemblies with and without friction have been thoroughly studied.  The computational complexity of determining the assembly stability in 2D is established in~\cite{palmer1989}. The stability determination techniques have been utilized in several robotic applications, that include a Jenga playing robot~\cite{wang2009}, multiple robots building a ramp~\cite{Napp2014}, an autonomous robot stacking a balancing vertical tower out of irregularly shaped stones~\cite{Furrer2017}, and a robot dry stacking irregular objects to build large piles~\cite{thangavelu2018}. Note that, in these studies, the challenging task planning aspect of construction planning has not been addressed.

{

Toussaint~\cite{toussaint2015} has utilized stability checks for building some tallest stable tower from a set of unlabeled cylinders and blocks; no goal condition is specified. His method applies a restricted version of task planning to decide for the order of manipulation actions, based on simple Strips operators and Monte Carlo tree search, and considers a restricted form of stability check that depends on whether the objects are placed on support areas of other objects. Due to these restrictions, his method is limited to building towers with sequential plans.

Note that for sophisticated constructions that involve temporary scaffolding, counterweights, and subassemblies, it is required to express ramifications of actions as well as true concurrency. However, expressing ramifications directly by simple Strips operators is not possible~\cite[Theorem~3]{thiebauxHN05} due to lack of logical inference. Also, expressing true concurrency is not possible unless the description is extended with exponential number of new operators, where each operator characterizes a concurrent action.  Due to these theoretical results, other studies~\cite{erdogan2013,erdogan2015,toussaint2017} that rely on simple Strips operators, do not present general methods for such sophisticated constructions either. 

It is important to note that these methods do not cover sophisticated structures, like bridges or overhangs, since objects are not necessarily placed on support areas of other objects. Such sophisticated structures require definition of transitive closure to ensure supportedness or connectedness. Transitive closure is not definable in first-order logic~\cite[Theorem~5]{Fagin75}; it is not directly supported by Strips either~\cite{thiebauxHN05}.

}

\textbf{Assembly planning:} In automated manufacturing, assembly plans aim to determine the proper order of assembly operations to build a coherent object. During assembly planning, the goal configuration is well-defined and the problem is generally approached by starting with the goal configuration and working backwards to disassemble all parts. Object stability has also been considered within this context~\cite{boneschanscher1988,lee1990,schmult1992,wilson1994,rohrdanz1996,beyeler2015,wan2018}.
The assembly planning problem is significantly different from the robotic construction problems: on the one hand, it allows assembly of irregular objects; on the other hand, the goal configuration is pre-determined and solutions are commonly restricted to monotone plans.

\textbf{Rearrangement planning:} Geometric rearrangement with multiple movable objects and its variations (like navigation among movable obstacles~\cite{stilman2005navigation,stilman2008planning}) have been studied in literature. Since even a simplified variant with only one movable obstacle has been proved to be NP-hard~\cite{wilfong1988,DemaineDHO03}, many studies introduce several important restrictions to the problem, like monotonicity of plans~\cite{cosgun2011push,dogar2012planning,stilman2007manipulation,barry2013manipulation,okada2004environment,KrontirisB16,KrontirisB15}. While a few can handle nonmonotone plans~\cite{havur2014,Krontiris2014}; these studies do not allow stacking either. Recently, Han et al.~\cite{HanSBY18} study rearrangement of objects in stack-like containers (by pushes and pops); these problems do not require stability checks.
%

%


\section{Robot Construction Problems}

The robot construction problem asks for a final stable configuration of different types of prefabricated blocks stacked on each other that satisfy some goal conditions, and a feasible stack rearrangement plan to obtain that final configuration from a specified initial configuration of the blocks. Figures~\ref{fig:overhang} and~\ref{fig:plans} present such stable final configurations, together with feasible construction plans to achieve them.

Initially, {
regular shaped boxes} are stacked on the ground/table as specified by the problem instance. The ground consists of a set of  surfaces (disconnected surfaces are required for bridges) and each surface possesses limited space for construction.

We use unit spaces, within a discrete model of the problem, to identify how much space is available on a box/surface and where to locate a box.  A single unit space is set to be equal to the size of the smallest box. To describe our approach, we consider three types of prefabricated blocks in the form of regular-shaped boxes: small boxes with one unit space, medium boxes with three unit spaces, and large boxes with five unit spaces. We assume that the width and the height of all the boxes are the same, while their weights may vary based on the problem instance.


We consider construction tasks performed by multiple autonomous robots, such as bimanual manipulators. The robots can pick and place boxes. We assume that the orientations of the boxes remain the same during the plan, so that the robots do not have to rotate the boxes. {
Our approach does not rely on any assumptions about the weight distribution of the boxes.} For clarity of presentation and without loss of generality, we only focus on the stability of the structures as the feasibility check performed in the continuous domain.  In particular, we ensure the stability of each step of the plan by testing it with a physics engine. Other feasibility checks, such as motion planning queries, reachability and graspability checks for manipulation actions, can be similarly integrated to our hybrid framework~\cite{ErdemPS16,Saribatur2018}.


The goal conditions may be described in an abstract manner to capture important aspects of specific structures. For instance, for a bridge, the ground on one side should be \emph{connected} to the one on the other side; for an overhang, constraints can be provided about the desired length of the overhang. If necessary, further goal conditions may be specified in an abstract manner (e.g., lightweight boxes should be placed on top of heavy ones), with more details (e.g., Boxes~3 and~4 must be placed on Box~5), or even with further details (e.g., Box~1 must be
placed on Box~2, ensuring that unit space~1 of Box~1 is on unit space~3 of Box~2).

In general, there exists multiple final configurations that satisfy the goal conditions, but only the ones that are stable and that can be achieved with a feasible construction plan are of interest. In that sense, the robot construction problem not only aims for a plan that reaches a goal configuration, but also ensures that this configuration and all intermediate steps are stable. Under these assumptions, we model the robot construction problems as a hybrid planning problem.


\section{Modeling the Robot Construction Problem}

We use Answer Set Programming (ASP)~\cite{BrewkaEL16}---a form of knowledge representation and reasoning paradigm in AI---for hybrid planning.  The idea is to represent the hybrid action domain by a set of logical formulas (called ``rules''), whose models (called ``answer sets''~\cite{gelfondL91}) correspond to plans and can be computed by special implemented systems called answer set solvers, like \dlvhex~\cite{dlvhex}, making calls to relevant feasibility checkers as needed.

\subsection{Formulas in ASP}

We consider disjunctive ASP rules of the form:
$$
\alpha_1 \vee \dots \vee \alpha_k  \leftarrow \beta_1, \dots,
\beta_n, \no\ \beta_{n+1}, \dots, \no\ \beta_m
$$
where $m,k \geq 0$, each $\alpha_i$ is an atom, and each $\beta_i$ is an atom or an external atom.
Intuitively, a rule expresses that if all $\beta_i$ ($1\leq i\leq n$) holds but no $\beta_i$ (${n+1}\leq i\leq m$) holds then some $\alpha_i$ ($1\leq i\leq k$) holds as well.
When $k=0$, the rule is a constraint; when $n=m=0$, it is a fact.


An external atom
$\ext{g}{Y_1,\dots,Y_n}{X_1,\dots,X_m}$ is defined by its name $g$, input $Y_1,\dots,Y_n$ and
output $X_1,\dots,X_m$. Intuitively, $g$ takes
the input $Y_1,\dots,Y_n$, passes it to an external computation (like a
stability checker), and conveys the results $X_1,\dots,X_m$ into the rules.

%

{
\subsection{Fluents and actions}

The objects in a robot construction world consist of a set $\cal A$ of robotic grippers, a set $\cal B$ of blocks, and a set $\cal L$ of locations (${\cal B}\subseteq {\cal L}$). The positions on each location $l\in {\cal L}$ (and thus each block $b\in{\cal B}$) is represented by its unit spaces $1..n_l$ for some positive integer $n_l$. Moreover, nonnegative integers $0..T-1$ describe time steps for a task plan, where $T$ specifies the maximum makespan (i.e., length) of a plan. In the following, the variable $t$ ranges between $0$ and $T$,
$a$ and $a'$ range over all grippers, $b$ and $b'$ range over all blocks, $l$ and $l'$ range over
all locations (e.g., blocks, table), and $u$, $u'$, $v$ and $v'$ range over relevant unit spaces.}

We consider two fluents to describe the
states of the world: $\ii{holding}(a,b,t)$ (robot's gripper $a$ is holding
box $b$ at step $t$ of the plan), and $on(b,l,u,v,t)$ (box $b$ is at location
$l$ at time step $t$, in such a way that the unit space $v$ of $b$ is on the
unit space $u$ on $l$).

We consider two actions: $\ii{pick}(a,b,t)$ (pick
the box $b$ with the gripper $a$ at step $t$) and $\ii{place}(a,l,t)$ (place
the box being held by the gripper $a$, on the location $l$ at step $t$) with
the attribute $\ii{placeOn}(a,b,l,u,v,t)$ (place the box $b$ being held by
the gripper $a$ such that the unit space $v$ of $b$ is on the unit space $u$ of
$l$). {
Here, the variable $t$ ranges between $0$ and $T{-}1$.}

{
Using these fluent and action constants, the preconditions and direct effects of these pick and place actions, the commonsense law of inertia, and the uniqueness and existence constraints for positions of unit spaces of blocks can be formalized in ASP following the guidelines described by Erdem et al.~\cite{ErdemP18,ErdemGL16}.
In the following, let us denote by $\Pi$ such an ASP program, and explain how the further challenges of robot construction problems are addressed using ASP. }

\subsection{Supportedness constraints}\label{sec:supported}

{
The supportedness constraints ensure that, at every state of the world, no box is supported by itself (i.e., no circular configurations). The formulation of these constraints is challenging because it requires the transitive closure of a binary relation \ii{onAux} that describes which block is on which location.  Fortunately, it is definable recursively in ASP.}

We obtain \ii{onAux} from \ii{on} by projection. For every step $t$, we recursively define supportedness of a block $b$ by a location $l$ as follows:
\beq
\ba l
\ii{supported}(b,l,t) \lar \ii{onAux}(b,l,t) \\
\ii{supported}(b,l,t) \lar \ii{onAux}(b,l',t), \\
\qquad \ii{supported}(l',l,t) \ \ (b{\neq} l') .
\ea
\eeq{eq:supported}

After that, we add a constraint to ensure that no box $b$ is supported by itself:
\beq
\lar \ii{supported}(b,b,t).
\eeq{eq:supported-cons}

{
\begin{prop}\label{prop:supported}
Given that the relative locations of blocks are described by fluents of the form $\ii{on}(l,b,u,v,t)$, rules (\ref{eq:supported}) correctly describe the supportedness of blocks by locations. Furthermore, adding rules (\ref{eq:supported}) $\cup$ (\ref{eq:supported-cons}) to $\Pi$ ensures that no circular configuration of blocks occur in construction at any time step $0..T-1$.
\end{prop} }

\smallskip
\noindent The proof follows from Theorem~2 and Proposition~5 of~\cite{ErdemL03} about the correctness and well-foundedness of the transitive closure of a relation defined recursively in ASP.

\subsection{Ramifications}\label{sec:ramifications}

The pick and place actions of the robot have many interesting indirect
effects (or ramifications). For instance, if a box $b$ is placed on some
location $l$, then as a direct effect of this action $b$ becomes on $l$; as
an indirect effect, the robot's gripper becomes empty.
Furthermore, if the unit space $v$ of box $b$ is on the unit space $u$ of location $l$, then
$(b,v)$ is not on any other unit $(l',u')$.

If a robot's gripper $a$ picks a box $b$, then as its direct effect $a$ is
holding $b$; as an indirect effect, $b$ is not on any box or the table.
Furthermore, as an  indirect effect, the gripper $a$ is not holding any
other box $b'$ ($b\neq b'$),
and no other gripper $a'$ is holding $b$ ($a\neq a'$).

\begin{figure}[htb]
\centering
\vspace{-.75\baselineskip}
\resizebox{0.6\columnwidth}{!}{\includegraphics{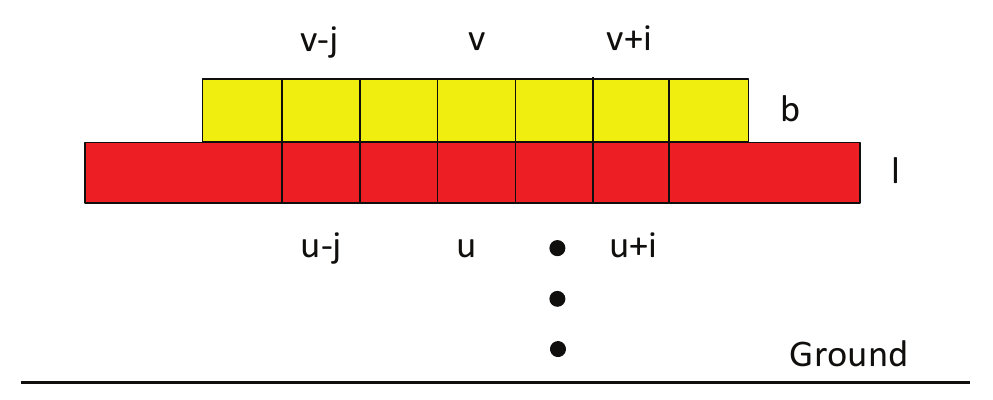}}
\vspace{-\baselineskip}
\caption{As indirect effects of placing unit $v$ of $b$ on unit $u$ of $l$, unit $v+i$ (resp. $v-j$) of $b$ is on unit $u+i$ (resp. $u-j$) of $l$.}
\vspace{-.5\baselineskip}
\label{fig:ramification1}
\end{figure}

An interesting ramification occurs when a longer box $b$ is placed on top of
another box: after the robot places the box $b$ being held its gripper $a$
onto the location $l$, so that unit space $v$ of $b$ is placed on which unit
$u$ of $l$, as an indirect effect the box $b$ occupies as many unit spaces as
its size allows on $l$.
We represent the ramifications of placing a longer box $b$ is on top of
another box $l$ as follows. Suppose that $b$ occupies the right part of $l$,
starting from the unit space $u$ of $l$ (Figure~\ref{fig:ramification1}). This can be expressed by the following
rule:
$$
\ii{on}(b,l,u+i,v+i,t) \lar \ii{on}(b,l,u,v,t)
$$
where $i$ ranges between $1$ and $\ii{min}\{\ii{size}(b)-v,\ii{size}(l)-u\}$, and $\ii{size}(b)$ denotes the length of the $b$.
Similarly, a rule is added for the left part of $l$ being occupied by $b$.

\begin{figure}[htb]
\centering
\resizebox{0.8\columnwidth}{!}{\includegraphics{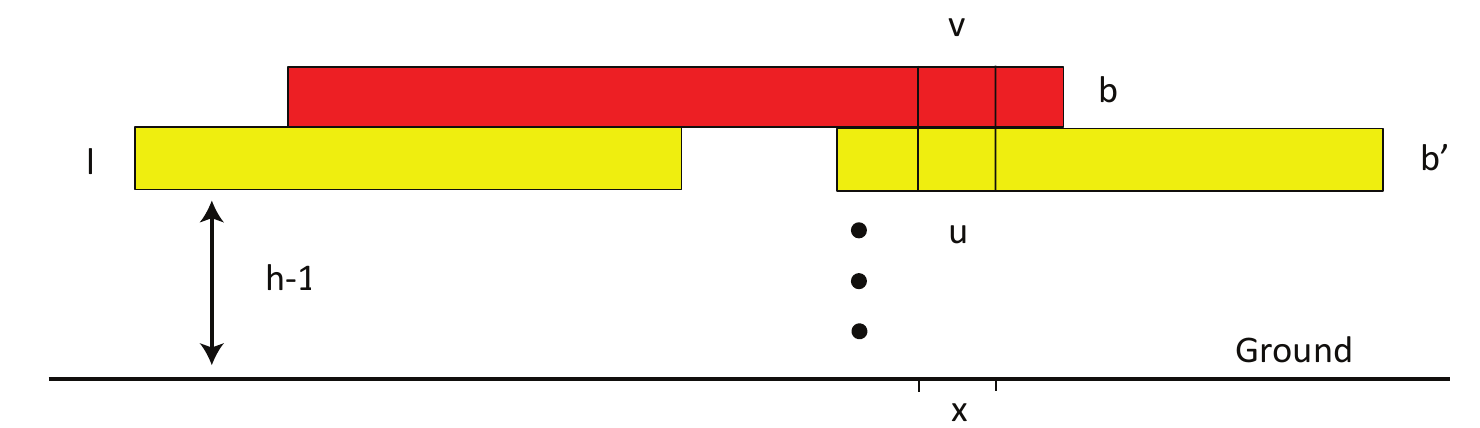}}
\vspace{-1\baselineskip}
\caption{As indirect effects of placing $b$ on $l$, unit $v$ of $b$ becomes on unit $u$ of $b'$, and unit $v+i$ (resp. $v-j$) of $b$ becomes on unit $u+i$ (resp. $u-j$) of $b'$.}
\vspace{-\baselineskip}
\label{fig:ramification2}
\end{figure}

Another interesting ramification occurs when a longer box $b$ is placed on top of
another box, but as a ramification it is also placed on a neighboring box $b'$
that is not too far (Figure~\ref{fig:ramification2}).
Such a sophisticated ramification is represented as follows.
First, we introduce some auxiliary atoms of the form $\ii{above}(h,b,v,x,t)$ to define the global locations of the boxes on the table, taking the leftmost side of the table as a reference; $\ii{above}(h,b,v,x,t)$ expresses that the unit space $v$ of the box $b$ at time step $t$ is at a global location that is $x$ units to the right of the leftmost side of the table and at $h$ units high from the surface of the table.
This predicate is defined with double recursion. The first recursion defines the global location of unit $v$ of a box $b$ vertically within a tower that is located $x$ units to the right of the leftmost side of the table:
\beq
\ba l
\ii{above}(1,b,v,x,t) \lar \ii{on}(b,\ii{Table},x,v,t) \\
\ii{above}(h,b,v,x,t) \lar \ii{above}(h{-}1,b',u,x,t), \\
\qquad \ii{on}(b,b',u,v,t) .
\ea
\eeq{eq:above1}
{
Note that this recursive definition characterizes, for every unit space $x$ of the table, the transitive closure of a binary relation that expresses which block unit $(b,v)$ is on which block unit $(b',u)$ over the unit space $x$ of the table.}
The second recursion defines the locations of other units of box $b$ horizontally to the right and to the left of that tower:
\beq
\ba l
\ii{above}(h,b,v{+}1,x{+}1,t) \lar \\
\qquad \ii{above}(h,b,v,x,t) \quad (v {<} size(b)) \\
\ii{above}(h,b,v{-}1,x{-}1,t) \lar \\
\qquad \ii{above}(h,b,v,x,t) \quad (v {>} 1)
\ea
\eeq{eq:above2}
{
This recursive definition also characterizes a set of transitive closures (i.e., for every height $h$, transitive closure of the adjacency relation of block units at the same height $h$).}

\begin{prop}\label{prop:above}
Given that the relative locations of blocks are described by fluents of the form $\ii{on}(l,b,u,v,t)$,
rules (\ref{eq:above1}) $\cup$ (\ref{eq:above2}) correctly describe the global positions of blocks.
\end{prop}

\smallskip
\noindent The proof follows from Theorem~2 of~\cite{ErdemL03} about the correctness of the transitive closure of a relation.

After that, using this auxiliary atom, we can represent the ramifications caused by placing a long box $b$ on top of a box, which is a horizontal-neighbor of another box $b'$: If we know that the unit space $v$ of $b$ and the unit space $u$ of $b'$ are both globally located horizontally $x$ units from the leftmost side of the table, and $b$ is at level $h$ and $b'$ is at level $h-1$, then the unit $v$ of $b$ is on the unit $u$ of $b'$.
$$
\ba l
\ii{on}(b,b',u,v,t) \lar \ii{above}(h,b,v,x,t), \ii{above}(h{-}1,b',u,x,t) \\
\ea
$$

\begin{figure*}[t]
	\centering
    \resizebox{1.75\columnwidth}{!}{\includegraphics{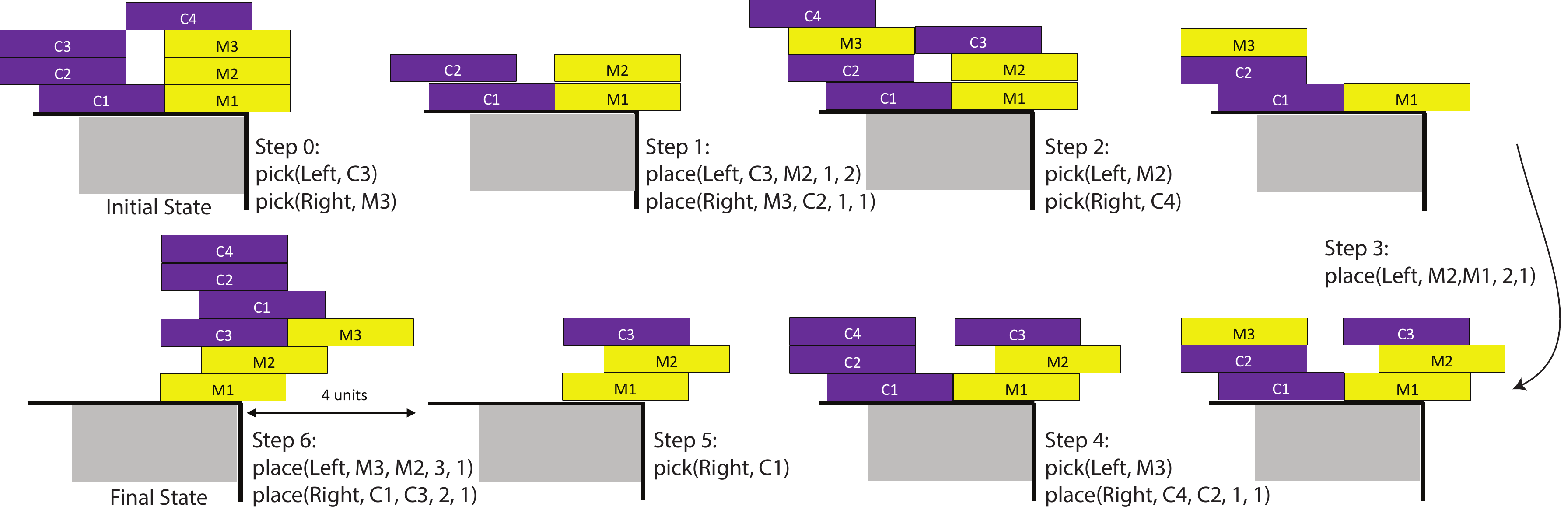}}
    \vspace{-1\baselineskip}
	\caption{A sample plan for a stable overhang construction}
    \vspace{-1\baselineskip}
	\label{fig:overhang}
\end{figure*}

\subsection{Stability check}\label{sec:stability}

{
Stability check ensures the stability of the overall assembly and each subassembly at every state of the construction plan. Stability checks are performed in a separate module and results of these checks are embedded in the ASP  formulation using external atoms. This modular approach enables our framework to be independent from any particular implementation of the stability checking algorithm, thus the stability checker can be treated as a black-box. Note that same approach is commonly employed for collision-checking during motion planning~{\cite{lavalle2006}}.


Let $\Gamma$ be a stability checking algorithm that returns \ii{True} if the given structure is stable, and \ii{False} otherwise.

We consider two external atoms to embed stability checks into our ASP formulation: $\ext{stable}{on,t}{}$ gets as input the relative positions of all the boxes supported by the table at step $t$ (described by the \ii{on} predicate); and $\ext{hStable}{holding,on,t}{}$ gets as input the relative positions of all the boxes being carried by a manipulator at step $t$ (described by $\ii{holding}$ and \ii{on} predicates). Both external atoms utilize the stability checker $\Gamma$, and return the outputs accordingly.

We embed the outcomes of stability checks into our domain description by constraints as follows:}
\beq
\ba l
\lar \no\ \ext{stable}{\ii{on},t}{} \\
\lar \ii{holding}(a,b,t), \ii{onAux}(b',b,t),\\
\quad 	\no \ \ext{hStable}{\ii{holding},\ii{on},t}{}.
\ea
\eeq{eq:stable}

{
\begin{prop}\label{prop:stability}
Let $\Pi'$ be the ASP program obtained from $\Pi$ by adding the supportedness constraints and the ramification rules as described above. Suppose that the stability checking algorithm $\Gamma$ is correct (i.e., the construction is stable iff $\Gamma$ returns \ii{True}).
Then adding rules (\ref{eq:stable}) to $\Pi'$ ensures that every configuration of blocks assembled on a flat surface (e.g., table) or being carried by a gripper at every time step $0..T-1$ during construction is stable.
\end{prop}

\smallskip
\noindent Propositions~\ref{prop:supported} and~\ref{prop:above} ensure constructions that satisfy supportedness constraints. Then the proof follows from application of Proposition~2 of~{\cite{Erdogan2004}} about the elimination of models by adding constraints.
}

\subsection{Concurrency constraints}\label{sec:concurrency}

Unless specified otherwise, the ASP modeling of the construction problem
allows true concurrency. We can specify noconcurrency constraints
explicitly. For instance, the concurrency of two pick actions of the same box
but with different grippers is not allowed with the following formula:
$$\lar \#\ii{count}\{a: \ii{arm}(a), \ii{pick}(a,b,t)\}>1 .$$


\subsection{Connectedness constraints}\label{sec:connected}

In bridge construction scenarios, one of the required conditions about a final structure is that there exists a block $x$ on the left side of the bridge and another block $y$ on the right hand side of the bridge such that $x$ and $y$ are connected to each other. For this reason, we recursively define connectedness of blocks using an auxiliary atom of the form $\ii{connected}(x,y,t)$ (block $x$ is supported by block $y$, or vice versa) similar to the recursive definitions that we have seen above, and add a constraint to express the required condition above for the goal (i.e., last time $T$):
\beq
\ba l
\lar \#\ii{count}\{x,y: \ii{connected}(x,y,T), \\
\qquad \ii{side}(x,\ii{Left},T), \ii{side}(y,\ii{Right},T)\}=0 .
\ea
\eeq{eq:connected}


{
\begin{prop}\label{prop:connected}
Suppose that the stability checking algorithm $\Gamma$ is correct (i.e., the construction is stable iff $\Gamma$ returns \ii{True}). Let $\Pi'$ be the ASP program obtained from $\Pi$ by adding the supportedness and stability constraints, and the ramification rules as described above.
Given that the supportedness of blocks and its inverse relation are described by atoms of the form $\ii{connected}(x,y,t)$,
adding constraints (\ref{eq:connected}) to $\Pi'$ ensures a stable bridge that connects two regions specified by \ii{Left} and \ii{Right}.
\end{prop}

\smallskip
\noindent Propositions~\ref{prop:supported}--\ref{prop:stability} ensure constructions that satisfy supportedness and stability constraints.  Then the proof follows from Proposition~3 and~2 of~\cite{Erdogan2004} about the conservative extensions of models by adding a definition, and elimination of models by adding constraints, respectively.
}

{
\subsection{Soundness and completeness}\label{sec:theory}

The soundness of the proposed method with respect to the desired properties is provided by Propositions~\ref{prop:supported}--\ref{prop:connected}.
The following proposition shows its completeness.

\begin{prop}\label{prop:complete}
Suppose that the stability checking algorithm $\Gamma$ is correct (i.e., the construction is stable iff $\Gamma$ returns \ii{True}).  Let $\Pi'$ be the ASP program obtained from $\Pi$ by adding the supportedness and stability constraints (and, in case of bridge construction, also the connectedness constraints), and the ramification rules as described above.
Then every robot construction plan that satisfies these desired properties and whose makespan is at most $T{-}1$ is characterized by an answer set for $\Pi'$.
\end{prop}

\smallskip
\noindent The proof follows from the representation methodology of the program: no constraint added to the program eliminates a valid robot construction plan.
}

\begin{figure*}[t]
	\centering
    \resizebox{1.65\columnwidth}{!}{\includegraphics{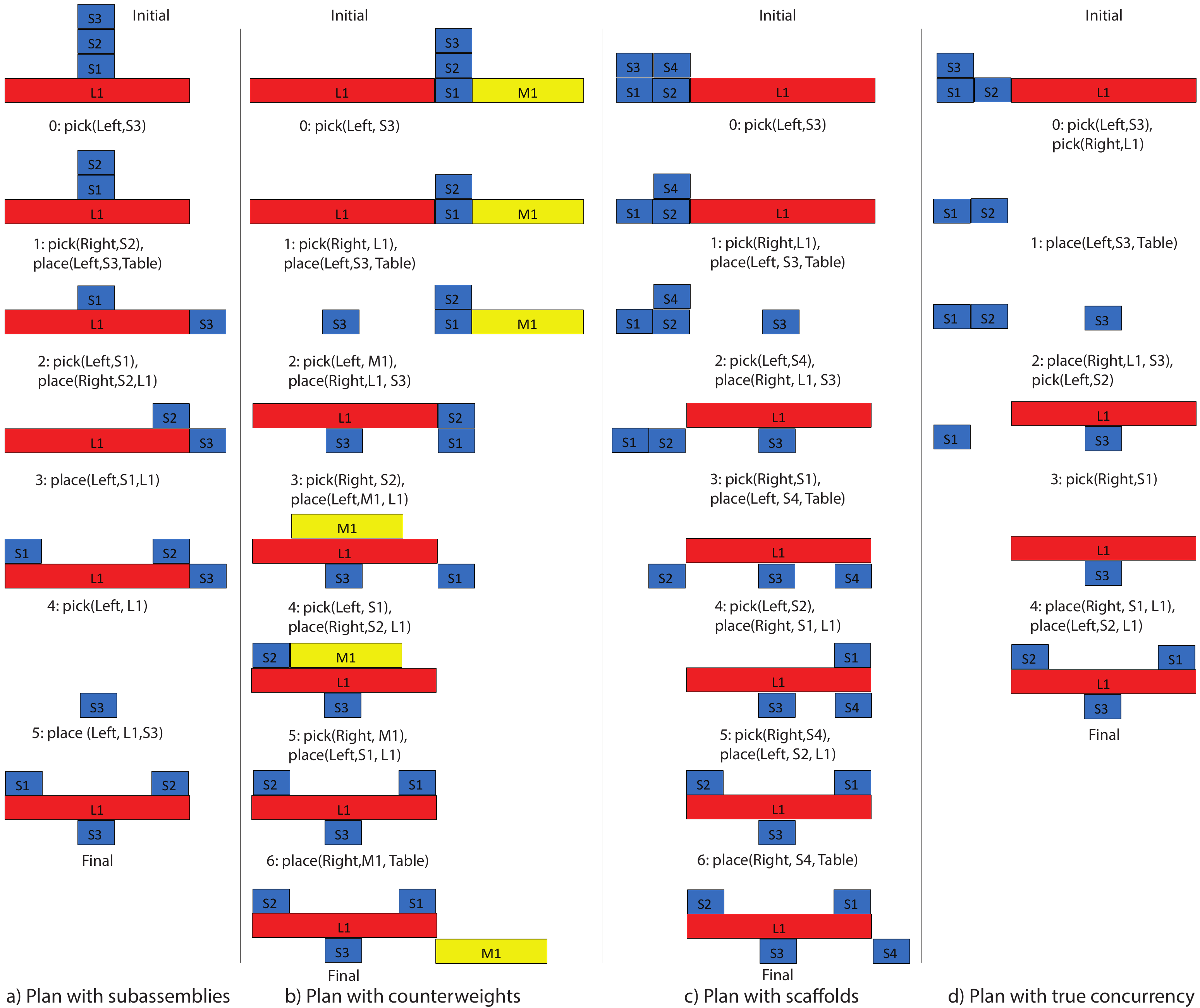}}
    \vspace{-.75\baselineskip}
	\caption{Plans for four challenging robot construction scenarios}
    \vspace{-1\baselineskip}
	\label{fig:plans}
\end{figure*}

\section{Sample Scenarios}

{
In our implementation, we test for stability numerically using the Pybullet physics engine. To ensure robust stability of our assemblies even under bounded disturbances, we adapt the notion of dynamic stability, in the sense of Fourier's inequality, which requires all objects to assume zero acceleration within a local neighbourhood their initial configuration, under the action of gravitational and friction forces~\cite{pang2000}. In addition to gravitational forces, we consider small disturbance forces to the assembly and check for its configuration after some finite time interval. If configuration of each object in the assembly stay within a empirically determined threshold from their initial location, we consider the assembly as dynamically stable.}

Figures~\ref{fig:overhang} and~\ref{fig:plans}  present several interesting scenarios to show the applicability of our formal approach to address challenging robot construction problems. In particular, Figure~\ref{fig:overhang} depicts a plan to construct an overhang of 4 units by a bimanual manipulator. Please note that the computed hybrid plan is not only nonmonotone, but also manipulates subassemblies and makes concurrent use of both arms.

Figure~\ref{fig:plans} demonstrates four scenarios that demonstrate the need for (a) manipulation of subassemblies, (b) utilization of counterweights, (c) use of scaffolds, and (d) true (non-serializable) concurrency to ensure stable construction of certain structures. Due to space limitations, feasible construction plans for the other interesting scenarios presented in Figures~\ref{fig:unlevel-bridges}--\ref{fig:bridges} are provided as a supplementary material at the project website.\footnote{~\url{http://cogrobo.sabanciuniv.edu/?p=1111}}




To verify the executability of the plans computed by our method, we have completed dynamic simulations and physical implementations of several benchmark scenarios using a Baxter robot. Implementation videos for these scenarios are available at the project website. 
%

\section{Conclusion}
{
We study multi-robot construction problems that are challenging for both AI and Robotics not only due to modeling challenges (e.g., due to ramifications of manipulation actions, true concurrency of actions, supportedness of blocks by other blocks), but also due to necessity of stability checks of constructions as they are being built. We address these challenges by a general hybrid planning framework developed over the logic-based formalism and automated reasoners of  Answer Set Programming (ASP): ASP allows true concurrency, embedding outcomes of stability checks into state constraints by external atoms, recursive definitions of sophisticated concepts, like supportedness and connectedness, and nested recursive definitions of global positions of blocks from their relative positions.

Our formal framework provides theoretical guarantees on soundness and completeness  with respect to desired properties of constructions (e.g., absence of nonsensical structures such as circular configurations of blocks, connectedness of the two sides of a bridge, stability of constructions).

We introduce a set of challenging robot construction benchmark instances that include bridges and overhangs constructed with counterweights, scaffolding and true concurrency of manipulations. Such a benchmark set of different types of construction is useful to advance studies on robot construction problems.

We demonstrate applications of our method over these benchmark instances. To the best of authors' knowledge, this is the first method to solve such a variety of challenging robot construction planning problems and only formal method with soundness and completeness guarantees.}

\bibliographystyle{IEEEtran}

\begin{thebibliography}{10}
\providecommand{\url}[1]{#1}
\csname url@samestyle\endcsname
\providecommand{\newblock}{\relax}
\providecommand{\bibinfo}[2]{#2}
\providecommand{\BIBentrySTDinterwordspacing}{\spaceskip=0pt\relax}
\providecommand{\BIBentryALTinterwordstretchfactor}{4}
\providecommand{\BIBentryALTinterwordspacing}{\spaceskip=\fontdimen2\font plus
\BIBentryALTinterwordstretchfactor\fontdimen3\font minus
  \fontdimen4\font\relax}
\providecommand{\BIBforeignlanguage}[2]{{%
\expandafter\ifx\csname l@#1\endcsname\relax
\typeout{** WARNING: IEEEtran.bst: No hyphenation pattern has been}%
\typeout{** loaded for the language `#1'. Using the pattern for}%
\typeout{** the default language instead.}%
\else
\language=\csname l@#1\endcsname
\fi
#2}}
\providecommand{\BIBdecl}{\relax}
\BIBdecl

\bibitem{bock2008}
T.~Bock, ``Construction automation and robotics,'' in \emph{Robotics and
  Automation in Construction}.\hskip 1em plus 0.5em minus 0.4em\relax InTech,
  2008.

\bibitem{WINOGRAD1972}
T.~Winograd, ``Understanding natural language,'' \emph{Cognitive Psychology},
  vol.~3, no.~1, pp. 1 -- 191, 1972.

\bibitem{gupta1992}
N.~Gupta and D.~S. Nau, ``On the complexity of blocks-world planning,''
  \emph{Artificial Intelligence}, vol.~56, no. 2-3, pp. 223--254, 1992.

\bibitem{fahlman1974}
S.~E. Fahlman, ``A planning system for robot construction tasks,''
  \emph{Artificial intelligence}, vol.~5, no.~1, pp. 1--49, 1974.

\bibitem{hall2005}
J.~F. Hall, ``Fun with stacking blocks,'' \emph{American journal of physics},
  vol.~73, no.~12, pp. 1107--1116, 2005.

\bibitem{paterson2006}
M.~Paterson and U.~Zwick, ``Overhang,'' in \emph{ACM-SIAM symposium on Discrete
  algorithm}, 2006, pp. 231--240.

\bibitem{paterson2009a}
M.~Paterson, Y.~Peres, M.~Thorup, P.~Winkler, and U.~Zwick, ``Maximum
  overhang,'' \emph{The American Mathematical Monthly}, vol. 116, no.~9, pp.
  763--787, 2009.

\bibitem{paterson2009b}
M.~Paterson and U.~Zwick, ``Overhang,'' \emph{The American Mathematical
  Monthly}, vol. 116, no.~1, pp. 19--44, 2009.

\bibitem{jia2015}
Z.~Jia, A.~C. Gallagher, A.~Saxena, and T.~Chen, ``3d reasoning from blocks to
  stability,'' \emph{IEEE Transactions on Pattern Analysis and Machine
  Intelligence}, vol.~37, no.~5, pp. 905--918, 2015.

\bibitem{stephenson2016}
M.~Stephenson and J.~Renz, ``Procedural generation of complex stable structures
  for angry birds levels,'' in \emph{IEEE Conference on Computational
  Intelligence and Games}, 2016, pp. 1--8.

\bibitem{walega2016}
P.~A. Wa{\l}ega, M.~Zawidzki, and T.~Lechowski, ``Qualitative physics in angry
  birds,'' \emph{IEEE Transactions on Computational Intelligence and AI in
  Games}, vol.~8, no.~2, pp. 152--165, 2016.

\bibitem{ferreira2014generating}
L.~Ferreira and C.~Toledo, ``Generating levels for physics-based puzzle games
  with estimation of distribution algorithms,'' in \emph{ACM Conference on
  Advances in Computer Entertainment Technology}, 2014, p.~25.

\bibitem{stephenson2016procedural}
M.~Stephenson and J.~Renz, ``Procedural generation of complex stable structures
  for angry birds levels,'' in \emph{Proc. of CIG}, 2016, pp. 1--8.

\bibitem{CalimeriFGHIR0T16}
F.~Calimeri, M.~Fink, S.~Germano, A.~Humenberger, G.~Ianni, C.~Redl,
  D.~Stepanova, A.~Tucci, and A.~Wimmer, ``Angry-hex: An artificial player for
  angry birds based on declarative knowledge bases,'' \emph{{IEEE} Trans.
  Comput. Intellig. and {AI} in Games}, vol.~8, no.~2, pp. 128--139, 2016.

\bibitem{blum1970}
M.~Blum, A.~Griffith, and B.~Neumann, ``A stability test for configurations of
  blocks,'' \emph{AIM}, 1970.

\bibitem{livesley1978}
R.~K. Livesley, ``Limit analysis of structures formed from rigid blocks,''
  \emph{International Journal for Numerical Methods in Engineering}, vol.~12,
  no.~12, pp. 1853--1871, 1978.

\bibitem{livesley1992}
------, ``A computational model for the limit analysis of three-dimensional
  masonry structures,'' \emph{Meccanica}, vol.~27, no.~3, pp. 161--172, 1992.

\bibitem{schimmels1994}
J.~M. Schimmels and M.~A. Peshkin, ``Force-assembly with friction,'' \emph{IEEE
  Transactions on Robotics and Automation}, vol.~10, no.~4, pp. 465--479, 1994.

\bibitem{mattikalli1996}
R.~Mattikalli, D.~Baraff, and P.~Khosla, ``Finding all stable orientations of
  assemblies with friction,'' \emph{IEEE Transactions on Robotics and
  Automation}, vol.~12, no.~2, pp. 290--301, 1996.

\bibitem{zwick2002}
U.~Zwick, ``Jenga,'' in \emph{ACM-SIAM Symposium on Discrete Algorithms}, 2002,
  pp. 243--246.

\bibitem{whiting2012}
E.~J.~W. Whiting, ``Design of structurally-sound masonry buildings using 3d
  static analysis,'' Ph.D. dissertation, Massachusetts Institute of Technology,
  2012.

\bibitem{mojtahedzadeh2015}
R.~Mojtahedzadeh, A.~Bouguerra, E.~Schaffernicht, and A.~J. Lilienthal,
  ``Support relation analysis and decision making for safe robotic manipulation
  tasks,'' \emph{Robotics and Autonomous Systems}, vol.~71, pp. 99--117, 2015.

\bibitem{pang2000}
J.-S. Pang and J.~Trinkle, ``Stability characterizations of rigid body contact
  problems with coulomb friction,'' \emph{ZAMM-Journal of Applied Mathematics
  and Mechanics/Zeitschrift f{\"u}r Angewandte Mathematik und Mechanik},
  vol.~80, no.~10, pp. 643--663, 2000.

\bibitem{spanos2001}
P.~D. Spanos, P.~C. Roussis, and N.~P. Politis, ``Dynamic analysis of stacked
  rigid blocks,'' \emph{Soil Dynamics and Earthquake Engineering}, vol.~21,
  no.~7, pp. 559--578, 2001.

\bibitem{palmer1989}
R.~S. Palmer, ``Computational complexity of motion and stability of polygons,''
  Cornell University, Tech. Rep., 1989.

\bibitem{wang2009}
J.~Wang, P.~Rogers, L.~Parker, D.~Brooks, and M.~Stilman, ``Robot jenga:
  Autonomous and strategic block extraction,'' in \emph{IEEE/RSJ International
  Conference on Intelligent Robots and Systems}, 2009, pp. 5248--5253.

\bibitem{Napp2014}
N.~Napp and R.~Nagpal, ``Distributed amorphous ramp construction in
  unstructured environments,'' \emph{Robotica}, vol.~32, no.~2, pp. 279--290,
  2014.

\bibitem{Furrer2017}
F.~Furrer, M.~Wermelinger, H.~Yoshida, F.~Gramazio, M.~Kohler, R.~Siegwart, and
  M.~Hutter, ``Autonomous robotic stone stacking with online next best object
  target pose planning,'' in \emph{IEEE International Conference on Robotics
  and Automation}, 2017, pp. 2350--2356.

\bibitem{thangavelu2018}
V.~Thangavelu, Y.~Liu, M.~Saboia, and N.~Napp, ``Dry stacking for automated
  construction with irregular objects,'' \emph{Artificial intelligence}, 2018.

\bibitem{toussaint2015}
M.~Toussaint, ``Logic-geometric programming: An optimization-based approach to
  combined task and motion planning.'' in \emph{IJCAI}, 2015, pp. 1930--1936.

\bibitem{thiebauxHN05}
S.~Thi{\'e}baux, J.~Hoffmann, and B.~Nebel, ``In defense of pddl axioms,''
  \emph{Artifiical Intelligence}, vol. 168, no. 1--2, pp. 38--69, 2005.

\bibitem{erdogan2013}
C.~Erdogan and M.~Stilman, ``Planning in constraint space: Automated design of
  functional structures,'' in \emph{IEEE International Conference on Robotics
  and Automation}.\hskip 1em plus 0.5em minus 0.4em\relax IEEE, 2013.

\bibitem{erdogan2015}
S.~D. Han, N.~M. Stiffler, K.~E. Bekris, and J.~Yu, ``Autonomous design of
  functional structures,'' \emph{Advanced Robotics}, vol.~29, no.~9, pp.
  625--638, 2015.

\bibitem{toussaint2017}
M.~Toussaint and M.~Lopes, ``Multi-bound tree search for logic-geometric
  programming in cooperative manipulation domains,'' in \emph{IEEE
  International Conference on Robotics and Automation}, 2017, pp. 4044--4051.

\bibitem{Fagin75}
R.~Fagin, ``Monadic generalized spectra,'' \emph{Mathematical Logic Quarterly},
  vol.~21, no.~1, pp. 89--96.

\bibitem{boneschanscher1988}
N.~Boneschanscher, H.~van~der Drift, S.~J. Buckley, and R.~H. Taylor,
  ``Subassembly stability,'' in \emph{AAAI}, vol.~88, 1988, pp. 780--785.

\bibitem{lee1990}
S.~Lee and Y.~G. Shin, ``Assembly planning based on geometric reasoning,''
  \emph{Computers \& graphics}, vol.~14, no.~2, pp. 237--250, 1990.

\bibitem{schmult1992}
B.~Schmult, ``Autonomous robotic disassembly in the blocks world,''
  \emph{International Journal of Robotics Research}, vol.~11, no.~5, pp.
  437--459, 1992.

\bibitem{wilson1994}
R.~H. Wilson and J.-C. Latombe, ``Geometric reasoning about mechanical
  assembly,'' \emph{Artificial Intelligence}, vol.~71, no.~2, pp. 371--396,
  1994.

\bibitem{rohrdanz1996}
F.~R{\"o}hrdanz, H.~Mosemann, and F.~Wahl, ``Generating and evaluating stable
  assembly sequences,'' \emph{Advanced robotics}, vol.~11, no.~2, pp. 97--126,
  1996.

\bibitem{beyeler2015}
L.~Beyeler, J.-C. Bazin, and E.~Whiting, ``A graph-based approach for discovery
  of stable deconstruction sequences,'' in \emph{Advances in Architectural
  Geometry}.\hskip 1em plus 0.5em minus 0.4em\relax Springer, 2015, pp.
  145--157.

\bibitem{wan2018}
W.~Wan, K.~Harada, and K.~Nagata, ``Assembly sequence planning for motion
  planning,'' \emph{Assembly Automation}, vol.~38, no.~2, pp. 195--206, 2018.

\bibitem{stilman2005navigation}
M.~Stilman and J.~J. Kuffner, ``Navigation among movable obstacles: Real-time
  reasoning in complex environments,'' \emph{International Journal of Humanoid
  Robotics}, vol.~2, no.~04, pp. 479--503, 2005.

\bibitem{stilman2008planning}
M.~Stilman and J.~Kuffner, ``Planning among movable obstacles with artificial
  constraints,'' \emph{The International Journal of Robotics Research},
  vol.~27, no. 11-12, pp. 1295--1307, 2008.

\bibitem{wilfong1988}
G.~Wilfong, ``Motion planning in the presence of movable obstacles,'' in
  \emph{Proceedings of the {F}ourth {A}nnual {S}ymposium on {C}omputational
  {G}eometry}, ser. SCG '88, 1988, pp. 279--288.

\bibitem{DemaineDHO03}
E.~D. Demaine, M.~L. Demaine, M.~Hoffmann, and J.~O'Rourke, ``Pushing blocks is
  hard,'' \emph{Comput. Geom.}, vol.~26, no.~1, pp. 21--36, 2003.

\bibitem{cosgun2011push}
A.~Cosgun, T.~Hermans, V.~Emeli, and M.~Stilman, ``Push planning for object
  placement on cluttered table surfaces,'' in \emph{IEEE/RSJ International
  Conference on Intelligent Robots and Systems}, 2011, pp. 4627--4632.

\bibitem{dogar2012planning}
M.~R. Dogar and S.~S. Srinivasa, ``A planning framework for non-prehensile
  manipulation under clutter and uncertainty,'' \emph{Autonomous Robots},
  vol.~33, no.~3, pp. 217--236, 2012.

\bibitem{stilman2007manipulation}
M.~Stilman, J.-U. Schamburek, J.~Kuffner, and T.~Asfour, ``Manipulation
  planning among movable obstacles,'' in \emph{Proceedings of ICRA 2007}.\hskip
  1em plus 0.5em minus 0.4em\relax IEEE, 2007, pp. 3327--3332.

\bibitem{barry2013manipulation}
J.~Barry, K.~Hsiao, L.~P. Kaelbling, and T.~Lozano-P{\'e}rez, ``Manipulation
  with multiple action types,'' in \emph{Experimental Robotics}.\hskip 1em plus
  0.5em minus 0.4em\relax Springer, 2013, pp. 531--545.

\bibitem{okada2004environment}
K.~Okada, A.~Haneda, H.~Nakai, M.~Inaba, and H.~Inoue, ``Environment
  manipulation planner for humanoid robots using task graph that generates
  action sequence,'' in \emph{IEEE IROS}, vol.~2, 2004, pp. 1174--1179.

\bibitem{KrontirisB16}
A.~Krontiris and K.~E. Bekris, ``Efficiently solving general rearrangement
  tasks: {A} fast extension primitive for an incremental sampling-based
  planner,'' in \emph{IEEE International Conference on Robotics and
  Automation}, 2016, pp. 3924--3931.

\bibitem{KrontirisB15}
------, ``Dealing with difficult instances of object rearrangement,'' in
  \emph{Robitics: Science and Systems}, 2015.

\bibitem{havur2014}
G.~Havur, G.~Ozbilgin, E.~Erdem, and V.~Patoglu, ``Geometric rearrangement of
  multiple movable objects on cluttered surfaces: {A} hybrid reasoning
  approach,'' in \emph{IEEE International Conference on Robotics and
  Automation}.\hskip 1em plus 0.5em minus 0.4em\relax IEEE, 2014, pp. 445--452.

\bibitem{Krontiris2014}
A.~Krontiris, R.~Shome, A.~Dobson, A.~Kimmel, and K.~Bekris, ``Rearranging
  similar objects with a manipulator using pebble graphs,'' in \emph{IEEE-RAS
  International Conference on Humanoid Robots}, 2014, pp. 1081--1087.

\bibitem{HanSBY18}
S.~D. Han, N.~M. Stiffler, K.~E. Bekris, and J.~Yu, ``Efficient, high-quality
  stack rearrangement,'' \emph{{IEEE} Robotics and Automation Letters}, vol.~3,
  no.~3, pp. 1608--1615, 2018.

\bibitem{ErdemPS16}
E.~Erdem, V.~Patoglu, and P.~Sch{\"{u}}ller, ``A systematic analysis of levels
  of integration between high-level task planning and low-level feasibility
  checks,'' \emph{{AI} Commununications}, vol.~29, no.~2, pp. 319--349, 2016.

\bibitem{Saribatur2018}
Z.~G. Saribatur, V.~Patoglu, and E.~Erdem, ``Finding optimal feasible global
  plans for multiple teams of heterogeneous robots using hybrid reasoning: an
  application to cognitive factories,'' \emph{Autonomous Robots}, pp. 1--26,
  2018, online first.

\bibitem{BrewkaEL16}
G.~Brewka, T.~Eiter, and M.~Truszczynski, ``Answer set programming: An
  introduction to the special issue,'' \emph{{AI} Magazine}, vol.~37, no.~3,
  pp. 5--6, 2016.

\bibitem{gelfondL91}
M.~Gelfond and V.~Lifschitz, ``Classical negation in logic programs and
  disjunctive databases,'' \emph{New Generation Computing}, vol.~9, pp.
  365--385, 1991.

\bibitem{dlvhex}
\BIBentryALTinterwordspacing
(2013, Aug.) dlvhex. [Online]. Available:
  \url{http://www.kr.tuwien.ac.at/research/systems/dlvhex/}
\BIBentrySTDinterwordspacing

\bibitem{ErdemP18}
E.~Erdem and V.~Patoglu, ``Applications of {ASP} in robotics,'' \emph{{KI}},
  vol.~32, no. 2-3, pp. 143--149, 2018.

\bibitem{ErdemGL16}
E.~Erdem, M.~Gelfond, and N.~Leone, ``Applications of answer set programming,''
  \emph{{AI} Magazine}, vol.~37, no.~3, pp. 53--68, 2016.

\bibitem{ErdemL03}
E.~Erdem and V.~Lifschitz, ``Tight logic programs,'' \emph{{TPLP}}, vol.~3, no.
  4-5, pp. 499--518, 2003.

\bibitem{lavalle2006}
S.~M. LaValle, \emph{Planning algorithms}.\hskip 1em plus 0.5em minus
  0.4em\relax Cambridge university press, 2006.

\bibitem{Erdogan2004}
S.~T. Erdogan and V.~Lifschitz, ``Definitions in answer set programming,'' in
  \emph{Proc. of LPNMR}, 2004, pp. 114--126.

\end{thebibliography}


\end{document}